\UseRawInputEncoding
\documentclass[
twocolumn,
]{ceurart}

\usepackage{listings}
\usepackage{graphicx}
\usepackage{array}
\usepackage{colortbl}
\usepackage{multirow}
\usepackage{rotating}
\usepackage{caption}
\usepackage{subcaption}
\usepackage{enumitem}

\begin{document}

\copyrightclause{Copyright © 2023}

\conference{SafeAI: The AAAI's Workshop on Artificial Intelligence Safety,
  Feb 13--14, 2023, Washington, D.C., US}

\title{Towards Developing Safety Assurance Cases for Learning-Enabled Medical Cyber-Physical Systems}


\author{Maryam Bagheri}[%
orcid=0000-0001-9576-2478,
]
\cormark[1]

\author{Josephine Lamp}[%
orcid=0000-0002-4982-7768,
]

\author{Xugui Zhou}[%
orcid=0000-0002-3663-7447,
]

\author{Lu Feng}[%
orcid=0000-0002-4651-8441,
]

\author{Homa Alemzadeh}[%
orcid=0000-0001-5279-842X,
]
\address{University of Virginia, Charlottesville, VA, USA, \{ntb5gu, jl4rj, xugui, lu.feng, alemzadeh\}@virginia.edu}

\cortext[1]{Corresponding author.}

\begin{abstract}
Machine Learning (ML) technologies have been increasingly adopted in Medical Cyber-Physical Systems (MCPS) to enable smart healthcare. Assuring the safety and effectiveness of learning-enabled MCPS is challenging, as such systems must account for diverse patient profiles and physiological dynamics and handle operational uncertainties. 
In this paper, we develop a safety assurance case 
for ML controllers in learning-enabled MCPS, with an emphasis on establishing confidence in the ML-based predictions. 
We present the safety assurance case in detail for Artificial Pancreas Systems (APS) as a representative application of learning-enabled MCPS, and provide a detailed analysis by implementing a deep neural network for the prediction in APS. We check the sufficiency of the ML data and analyze the correctness of the ML-based prediction using formal verification. 
Finally, we outline open research problems based on our experience in this paper.
\end{abstract}


\maketitle

\section{Introduction}
\vspace{-0.5em}


Medical Cyber-Physical Systems (MCPS) integrate connected software and hardware components with sensors and actuators to monitor and control  
patient physiology. Machine Learning (ML) technologies have been increasingly used in MCPS, often deployed in the estimation and prediction components, to make data-driven decisions based on sensor or patient input and guide control actions~\cite{JOUR}. 
Ensuring the successful deployment of ML within MCPS can be challenging, as the ML components must be able to handle the intricacies of patient physiology, time lags between the impact of a control action and sensor measurements, uncertainties in the operational environment that may affect the patient's physiology, and variability in patient profiles which may result in differing impacts of the control actions. Moreover, due to physiological complexities and the limited availability of realistic patient profiles or datasets, ML techniques may use synthetic data or virtual patient models for training. The mismatch between the training data and the real-world data seen in deployment may result in erroneous, biased, or incomplete output predictions \cite{10.1145/3453444}.
%
%
Failure of the ML component due to any of the challenges noted above could result in irreparable harm to patients.  As such, the use of ML within MCPS should be assured by evidence that these components are safe and reliable \cite{10.1145/3453444}.  



Assurance cases (AC) are structured arguments, supported by evidence, to justify claims for an application in a given environment \cite{10.1007/978-1-84996-086-1_4}. Recent efforts in safety analysis and assurance have confirmed that AC are valuable for assessing and demonstrating trust \cite{9269875}. For example, AC have been deployed for unmanned aircraft systems \cite{clothier2017making} and medical devices \cite{feng_et_al:OASIcs:2014:4526}. The U.S. FDA has also issued a guideline \cite{FDAGuideline} suggesting medical manufacturers provide AC with pre-market submissions. 
Among the standards published by various organizations 
(e.g., ISO 26262 \cite{ISO26262}, ISO/PAS 21448 \cite{ISOPAS}, ANSI/UL 4600~\cite{ANSI/UL}, and the FDA guideline for Artificial Pancreas Systems (APS) \cite{FDAGAPS}), ANSI/UL 4600 is the only one offering evaluations of ML technology for the safety of autonomous vehicles. 

Among more AC-centric studies, Hawkins et al.~\cite{https://doi.org/10.48550/arxiv.2102.01564} provide a guideline on the assurance of ML in autonomous systems, including a safety case pattern for each stage of the ML life cycle. Kaur et al. \cite{10.1007/978-3-030-55583-2_6} proposed a modular assurance case pattern based on assume/guarantee reasoning for ML-enabled CPS, where the safety of the ML component and the rest of the system are assessed separately. However, the ML lifecycle is not dealt with in this pattern. A few studies have also developed AC for concrete learning-enabled use cases in the automotive domain \cite{10.1007/978-3-030-83903-1_10,10.1007/978-3-030-54549-9_13,10.1007/978-3-030-26250-1_30}.  Within the healthcare domain, \cite{10.1007/978-3-030-26601-1_12} is the only work that presents an assurance case pattern to justify the use of ML. Even so, none of~\cite{https://doi.org/10.48550/arxiv.2102.01564}-\cite{10.1007/978-3-030-26601-1_12} instantiate the process activities and generate evidence for a concrete application in the medical domain. This paper tackles this gap by presenting detailed AC to assure the safety and effectiveness of a general framework of APS \cite{kapil2020artificial}, suitable for all types of APS. We select APS as a representative of learning-enabled MCPS as it contains a collection of typical components of many ML-enabled MCPS, including data-driven estimation algorithms and embedded controllers for insulin dosage calculation.

Expanding on the patterns proposed in \cite{https://doi.org/10.48550/arxiv.2102.01564,10.1007/978-3-030-55583-2_6}, we develop a safety assurance case for the ML-enabled controller of MCPS, consisting of a controller algorithm and an ML prediction algorithm, where our emphasis is greatly on the ML-based prediction. 
By proving trust in ML prediction, the safety of the controller algorithm can be assessed in isolation. Considering the patient as an essential element of the control loop, we discuss the AC elements that should be instantiated based on an individual patient profile or a population of patients. The instantiation is due to different physiologies of different patients, which may affect control actions, the ML controller's expected behavior, and the claims satisfaction. 
This is the first time that the patient profiles are included in AC for ML controllers and MCPS. 
In support of claims in AC for APS, we implement a deep neural network for blood glucose prediction in APS. We then present an analysis characterizing the sufficiency of the training data for the ML controller and the ML development process. 
We also utilize APS domain knowledge to specify a set of properties based on a patient's metabolism (e.g., insulin senstivity, carbohydrate absorption profile) 
and use formal verification to check them against the ML prediction component. We are unaware of any work verifying the ML components in APS except~\cite{10.1145/3365365.3382210}, which unlike us, compares the output ranges of two identical networks given a slight change in their input ranges.

\noindent \textbf{Contributions.} The major contributions of this paper are summarized as follows:
\begin{itemize}[itemsep=0em,leftmargin=*]
\vspace{-0.4em}
    \item We present preliminary results on developing a safety assurance case template for ML controllers in MCPS, which includes patient profiles in its element descriptions.
    \item We present a detailed safety assurance case for APS that is supported by a thorough analysis of ML-based glucose prediction module.
    \item 
    We define properties based on the body's metabolism and check them against the ML prediction component using formal verification.
\end{itemize}
\vspace{-0.4em}
In the end, we discuss open research problems in developing safety assurance cases for learning-enabled MCPS.

\vspace{-1em}

\section{Artificial Pancreas Systems} \label{sec::background}
\vspace{-0.5em}

Type 1 Diabetes (T1D) is a chronic disease in which a patient's pancreas produces little to no insulin.  Patients with T1D must constantly monitor their blood glucose (BG) levels and inject insulin to regulate their concentrations of BG. 
\begin{figure}
    \centering
    \fbox{\includegraphics[width=0.95\linewidth,keepaspectratio]{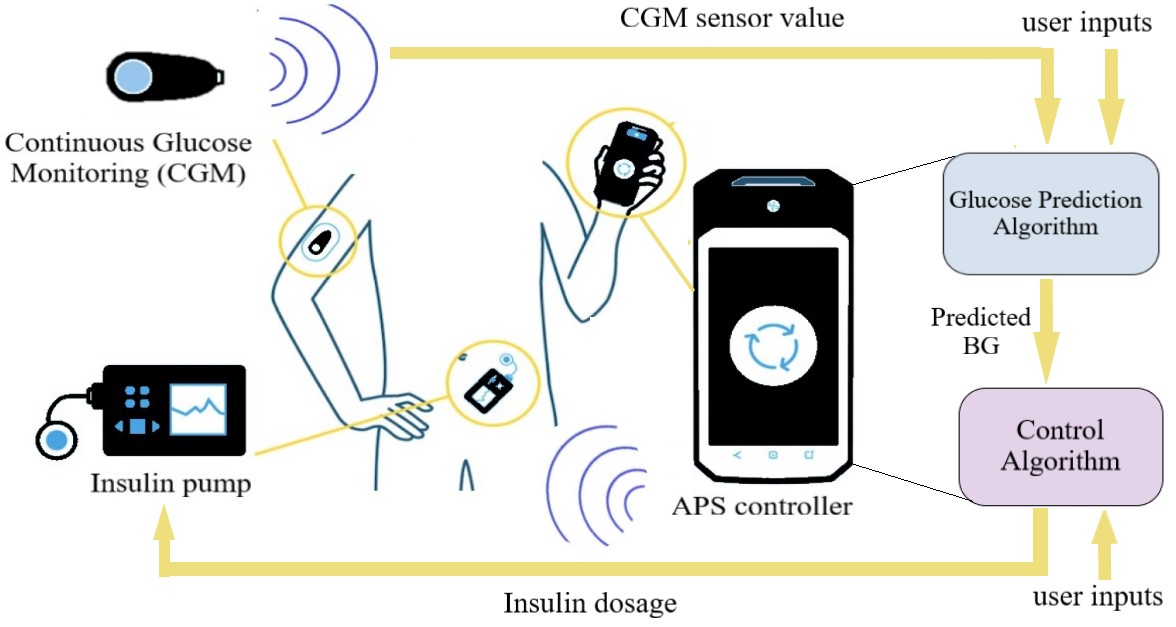}}
    \caption{The structure of  APS (modified from \cite{apsStructure}). The APS controller consists of a prediction algorithm to predict the BG values and a controller algorithm to adjust the insulin dosage.}
    \label{fig::apsstructure}
    \vspace{-2.5em}
\end{figure}
 Artificial Pancreas Systems (APS) are closed-loop insulin delivery systems that relieve the burden of T1D on patients by regulating a patient’s BG level, using input from various sensors such as continuous glucose monitors (CGM). Figure~\ref{fig::apsstructure} depicts the typical structure of APS, consisting of  a CGM sensor to continuously monitor BG values, an APS controller that calculates the correct insulin dosages based on the CGM and user input, and an insulin pump that delivers insulin dosages. 
 
 The APS controller consists of a data-driven glucose prediction algorithm to predict future BG values and a control algorithm to adjust the insulin dosage based on the predicted BG values. 
Recently, researchers have begun exploring the use of neural networks \cite{10.1007/978-3-319-99429-1_11,ZHANG2021102923} and reinforcement learning \cite{9115809} for the design of the APS controllers (e.g., use of machine learning for glucose prediction).  
The primary objective of the controller is to provide safe and efficient glycemic control by infusing an appropriate amount of insulin to keep the patient's BG within the proper range (between 70 and 180 mg/dL) and avoid hypoglycemias and hyperglcemias. 
To provide such safe control, the controller needs to account for the complexities of glucose metabolism and 
 deal with un-predicted meal intake, exercise, stress, or illness, rapid changes in BG concentration, and time lags between BG measurement and insulin impact.

As of the date of this writing, there are four commercially available APS that have received FDA approval and/or the Conformit\'{e} Europ\'{e}enne (CE) mark: Medtronic MiniMed 670/770/780G \cite{fdaminimed}, Tandem Control-IQ \cite{kovatchev2017feasibility}, Omnipod 5 \cite{sherr2020safety}, and CamAPS FX \cite{chen2021user}. 
These systems use some form of data-driven learning algorithm, i.e., 
Tandem Control-IQ uses a simple linear regression algorithm \cite{tandemAlgorithm} to predict BG values 30 minutes in the future and then a PID algorithm to adjust the insulin dosage based on the predicted BG values. To ensure the adoption and use of such systems, patients need to be confident in the underlying ML technology embedded in the controllers. 

\vspace{-1em}
\section{Overall Safety Assurance Case} \label{sec::overalFramework}
\vspace{-0.5em}
Typical CPS consist of embedded software and hardware components controlling the plant through interconnected sensors and actuators. MCPS are a distinct class of CPS with the patient as the plant, aiming to monitor and control multiple aspects of the patient's physiology. With the patient in the control loop, the MCPS should either be tailored for specific physiological parameters of the patient or should cover a population of patients. This adaptation is even more critical as a part of the design process in learning-enabled MCPS that employ a learning-enabled controller, a controller relying on machine learning to perform perception or prediction tasks. So, in the regulatory process for checking the safety of MCPS, it would make sense to instantiate the safety assurance case for individual patient profiles or populations. To emphasize this, we mark the context elements in safety assurance cases with the uppercase letter P in a half circle to denote the decision points where the context needs to be initialized for an individual patient profile or the population.

The top-level goal in safety AC of learning-enabled MCPS is to ensure that "x as a case of learning-enabled MCPS is safe and effective". Confidence in this claim is obtained by ensuring the safety and effectiveness of all constituent components, including sensors, actuators, and their interactions. For this paper, we discuss only the safety and effectiveness of the \textit{learning-enabled controller}, assuming that the safety and effectiveness of 
other components have been adequately examined. We first present a safety assurance case template for a learning-enabled controller in MCPS, shown in Figure~\ref{fig::overview}. We then use APS as an instance of MCPS and explain how instantiating the template results in a general safety assurance case for APS. We use goal structuring notation (GSN)~\cite{GSN}, with a slight of notation abuse, to show our AC. 


\vspace{-0.5em}
\subsection{Safety Assurance Case Template for learning-enabled controllers in MCPS}

\begin{figure}[t!]
    \centering
    \fbox{\includegraphics[width=0.9\linewidth,keepaspectratio]{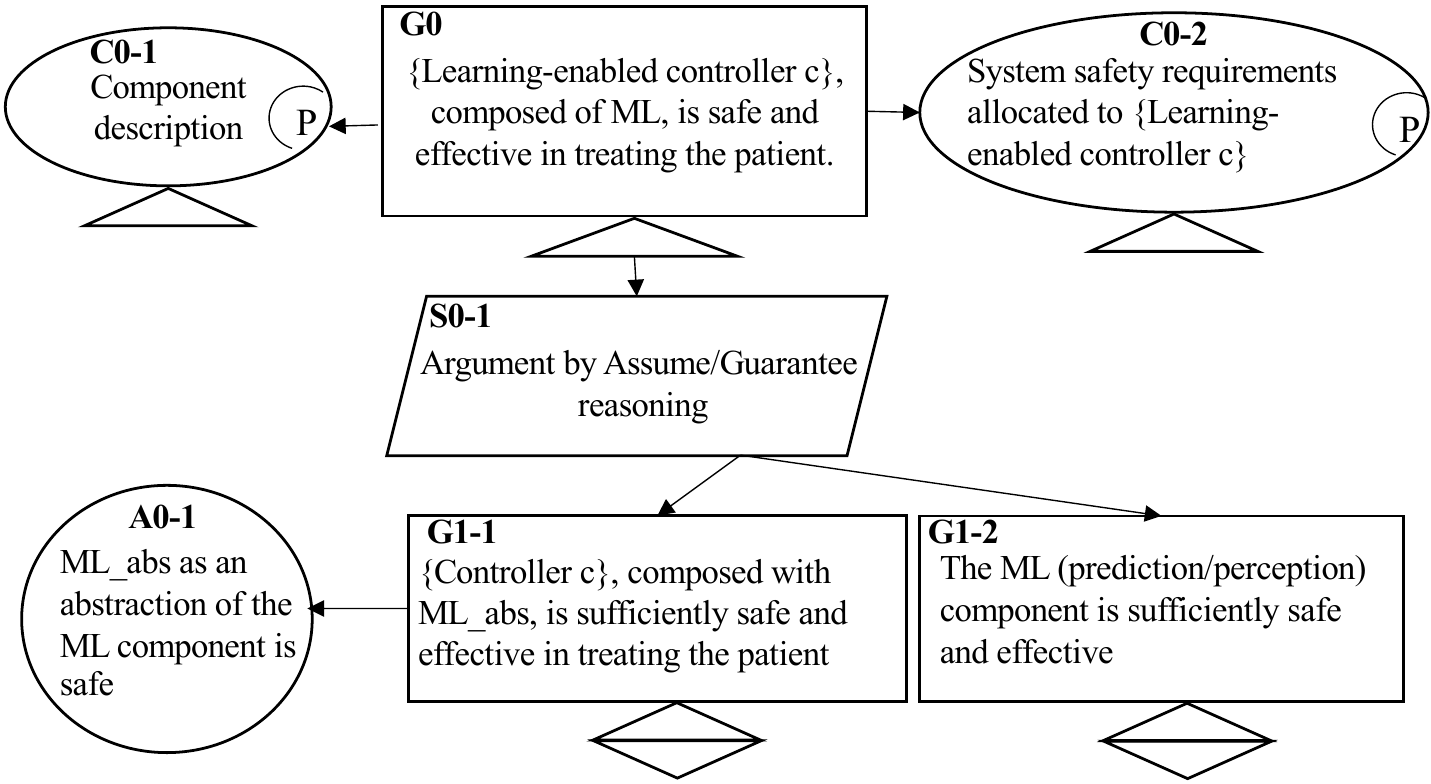}}
    \caption{A safety assurance case template for a learning-enabled controller in MCPS. 
    }
    \label{fig::overview}
    \vspace{-2em}
\end{figure}

The root goal G0 in Figure~\ref{fig::overview} asserts that the learning-enabled controller $c$ is safe and effective while the device is used in treating the patient. The environment and the system within which the learning-enabled controller is used are described in context C0-1, and the requirements assigned to the learning-enabled controller are explained in context C0-2. 
We use assume/guarantee reasoning to justify goal G0. This is because the learning-enabled controller $c$ is a combination of two main algorithms that perform in sequence: an algorithm that performs the ML tasks and delivers its output to the control algorithm, and the control algorithm that selects the control action and initiates it in the system. Assuming that the results provided by the ML tasks are correct, the control algorithm should guarantee the safe and effective treatment of the patient. Hereafter, we use the term controller to refer to the control algorithm of the learning-enabled controller. We consider a separate component for each algorithm. As reflected in goal G1-2, the safety and effectiveness of the ML component are justified in isolation. 
The controller's input is an interface with the ML component, containing the results received from the ML component. 
This interface shows an abstraction of the ML component, denoted as $ML\_abs$ in goal G1-1. Assuming the outputs of the ML component are safe, as presented in assumption A0-1, goal G1-1 claims that the controller component combined with $ML\_abs$ is safe and effective.

\vspace{-0.6em}
\subsection{Instantiating Learning-Enabled Controller Assurance Case for APS}
\begin{table*}[]
\footnotesize
\centering
    \caption{Requirements for the learning-enabled APS controller}
    
    \begin{tabular}{l|l}
        \hline 
         RQ.C.1 & Accurately calculate dose of basal and bolus insulin \\ \hline
         RQ.C.1.1 & Determine the output every T minutes (e.g., T=5 in MiniMed) \\ \hline
         RQ.C.1.2 & Stop dosing if a maximum amount has been delivered by the pump \\ \hline
         RQ.C.1.3 & Suspend dosing if the actual or predicted CGM readings fall below a threshold \\ \hline
         RQ.C.1.4 & Interrupt in a safe way if trustworthy control is not guaranteed \\ \hline
         RQ.C.1.5 & BG should not remain below 10th-percentile threshold for more than $\alpha_1$ minutes \\ \hline
         RQ.C.1.6 & BG should not remain above 90th-percentile threshold for more than $\alpha_2$ minutes following a bolus injection \\ \hline
         RQ.C.1.7 & BG should not remain above 90th-percentile threshold for more than $\alpha_3$ minutes \\ \hline
         RQ.C.1.8 & The BG value is always greater than 70 and less than 180 \\ \hline
         RQ.C.1.9 & The controller infuses additional insulin while the blood glucose level is below a target level \\ \hline
         RQ.C.1.10 & The morning wake up blood glucose level can not exceed $\beta$ \\ \hline
          \end{tabular}
    \label{tab::perRobReq2}
    \vspace{-2em}
\end{table*}

Regardless of the system type and the technology underlying the APS, the main components of APS remain the same, i.e., they all consist of a learning-enabled controller. Additionally, APS should satisfy a set of safety and performance properties common and desirable to regulatory agencies. For instance, requirements specified by the FDA \cite{FDAGAPS} include high accuracy of CGM readings, safe insulin dosages, usable design, and so on. 
The most significant requirement is that APS must not increase the incidence and severity of hypoglycemic and hyperglycemic events. These reasons induce a general safety assurance case to be developed for all APS, like~\cite{infusPump} which presents AC for a generic infusion pump device.   
Thus, the proposed template in Figure~\ref{fig::overview} can be employed for APS, where goals G1-1 and G1-2 are modified as follows. 

 \noindent  \textbullet\ G1-1: Assuming that the BG predictions are accurate, the insulin dosage management component is sufficiently safe and effective for treating patients. \\
    \noindent \textbullet\ G1-2: The ML glucose prediction component is sufficiently safe and effective.

\textbf{Context Elements}. Context C0-1 describes inputs to the APS controller (i.e., history of the CGM values and insulin injected  and prediction horizon), outputs of the controller (i.e., the amount of insulin to be injected), the component's role in the system, and the environmental phenomena (i.e., uncertain meal intake, daily activity).  

Context C0-2 includes all requirements of the learning-enabled controller. Table~\ref{tab::perRobReq2} shows a set of these requirements, which we have extracted from the diabetes treatment literature (e.g., \cite{10.1007/978-3-319-23820-3_1}). The main requirement in Table~\ref{tab::perRobReq2} is RQ.C.1, which is further refined into its following requirements. Although this is not an extensive list of requirements, it represents some of the most important requirements for such a system.  A few examples of peripheral requirements are related to the controller's platform, such as security, reliability, and usability. 
For example, the smartphone used in some of the APS is a platform. The goal G1-1 is also defined in the same context as C0-2 since it claims the safety and effectiveness of the APS controller when BG predictions are reliable. We assume that the controller  requirements are adequately examined in consultation with domain experts and medical professionals and do not discuss them in this paper. 

  



\textbf{Patient or Population}. The contexts C0-1 and C0-2 can be defined based on an individual patient profile or a population of patients. A clear example of this decision is that different patients have varying insulin sensitivity levels, and their physiologies may be affected differently by meal volume or activity. In addition to the training datasets and the control algorithms themselves, other components such as thresholds and target values that affect the requirements, can be defined differently based on individual patients or a population of patients. 

In the next section, we develop a general argument for goal G1-2, claiming that the ML glucose prediction component is sufficiently safe and effective. 

\vspace{-1em}
\section{Safety Assurance Case for the  Glucose Prediction Component} \label{sec::controllerArg}
\vspace{-0.5em}

\begin{table*}[t]
\footnotesize
    \caption{Performance and robustness requirements (that are independent of ML technology) for the ML glucose prediction component
    }
    \centering
    \begin{tabular}{m{2cm}|m{11cm}}
        \hline 
          \multicolumn{2}{|c|}{\textbf{Performance} } \\
          \hline \hline
         ML-RQ1 & Accurately predict the BG values $T$ minutes in the future \\ \hline
          ML-RQ1.1 & BG's rate of change has to be bound by established physiological norms \\ \hline
          ML-RQ1.2 & Meal intake has a direct effect on the BG value \\ \hline
          ML-RQ1.3 & Exercise has an inverse effect on the BG value  \\ \hline 
          ML-RQ1.4 & Within $t$ minutes of a bolus, there should be an accompanying change in BG of more than $\alpha$ \\ \hline
          ML-RQ1.5 &  The glucose level starts to rise at a specific time after a meal's onset \\ \hline        
          ML-RQ1.6 &  There is a delay between the injection of insulin and the disposal of glucose  \\ \hline   
          ML-RQ1.7 &  The blood concentration of insulin reaches its maximum after a particular time  \\ \hline  
          ML-RQ1.8 &  Insulin has an inverse effect on the BG value \\ \hline  
         \multicolumn{2}{|c|}{\textbf{Robustness}} \\ \hline \hline
          ML-RQ2 & Perform as required for different patients of different ages/sexes \\ \hline
          ML-RQ3 & Perform as required in the presence of external factors such as meals and exercises. \\\hline
    \end{tabular}
    \label{tab::perRobReq}
    \vspace{-2em}
\end{table*}
Different parts of the safety assurance case developed for the glucose prediction component are shown in Figures~\ref{fig::MLPartI}, \ref{fig::MLPartII}, and \ref{fig::MLPartIII}. We describe each part in a separate section and provide concrete pieces of evidence in Section~\ref{sec::concEvid}. 

\vspace{-1em}
\subsection{Sufficiency of the ML Development}
\begin{figure}
    \centering
    \fbox{\includegraphics[width=0.9\linewidth,keepaspectratio]{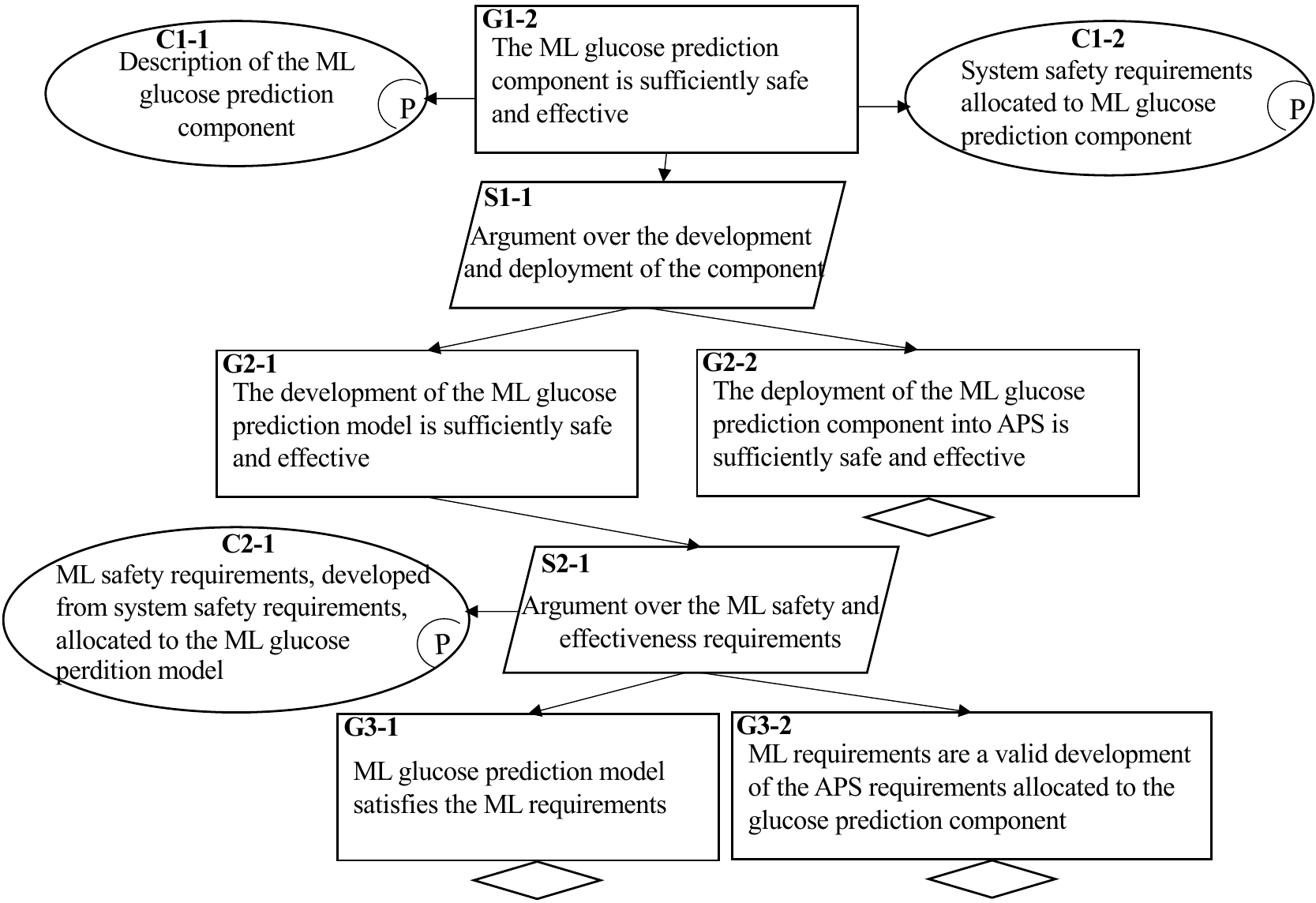}}
    \caption{A general safety assurance case for the ML glucose prediction component of APS. 
    }
    \label{fig::MLPartI}
    \vspace{-2.5em}
\end{figure}

The argument to justify claim G1-2 is shown in Figure~\ref{fig::MLPartI}. This claim is supported by contexts C1-1 and C1-2. Context C1-1 describes the ML prediction component, its expected inputs and outputs (e.g., CGM, insulin, and meal values), along with their possible sources and targets (e.g., various CGM or pump devices). Besides, it is necessary to determine whether the component is specialized based on a patient profile or a population of patients.  

We categorize the requirements allocated to the ML glucose prediction component into performance and robustness requirements and enumerate them in Table~\ref{tab::perRobReq}. These requirements are independent of ML technology and are defined in context C1-2. ML-RQ1 is the primary performance requirement refined into ML-RQ1.1 through ML-RQ1.8 based on the patient's physiology. Prediction results are accurate when the learning component has learned the physiological dynamics of the patients, and hence ML-RQ1.1 to ML-RQ1.8 are satisfied. Although we have extracted these requirements from the APS literature, based on our knowledge, it is the first time that physiologically-inspired requirements are assigned to an ML component. The thresholds and target values in these requirements depend on the patient's or population's profiles. 
Robustness requirements ML-RQ2 and ML-RQ3 refer to variations in the input space of the component. For instance, ML-RQ2 ensures a variety of patient profiles are considered in a population-based setting.  


%
To justify claim G1-2, the approach of \cite{https://doi.org/10.48550/arxiv.2102.01564} splits the argument based on the development and deployment of the ML component. Goal G2-1 claims that the development of the ML model predicting the BG values is sufficiently safe and effective, and goal G2-2 claims that the integration of the ML component into the system is sufficiently safe and effective. The G2-2 justification involves techniques such as runtime assessment that are beyond the scope of this paper. We leave G2-2 undeveloped and emphasize G2-1. The first step to support claim G2-1 is to develop ML requirements using the concepts amenable to the ML implementation. The performance requirement ML-RQ1 in Table~\ref{tab::perRobReq} can be measured by the accuracy or mean prediction error of the ML algorithm. Thus, ML-RQ1 is defined as `` ML component should predict the glucose value with the mean prediction error of less than \textit{thres} mg/dL'', where \textit{thres} is determined by human experts or compared to the most reliable existing method for BG prediction. ML-RQ1.1 to ML-RQ1.8 are meaningful to the ML model when defined over inputs and outputs of the ML model. We specify these requirements in Section~\ref{sec::concEvid}. 
These ML requirements are expressed in context C2-1.

The development of the ML component refers to the process of designing and training the ML model. So, claim G2-1 is supported by sub-claims G3-1 and G3-2 through strategy S2-1. Goal G3-1 claims that the ML model satisfies the ML requirements. A complete argumentation must demonstrate that the ML requirements are a valid development of the APS requirements allocated to the glucose prediction component, as expressed in claim G3-2. In our case, there is an exact mapping between requirements of Table~\ref{tab::perRobReq} and the ML requirements (Section~\ref{sec::concEvid}), so G3-2 is justified. 
The G3-1 justification is described in the next section.

\vspace{-1em}
\subsection{Sufficiency of the ML Model}
\vspace{-0.5em}

\begin{figure}
    \centering
    \fbox{\includegraphics[width=0.95\linewidth,keepaspectratio]{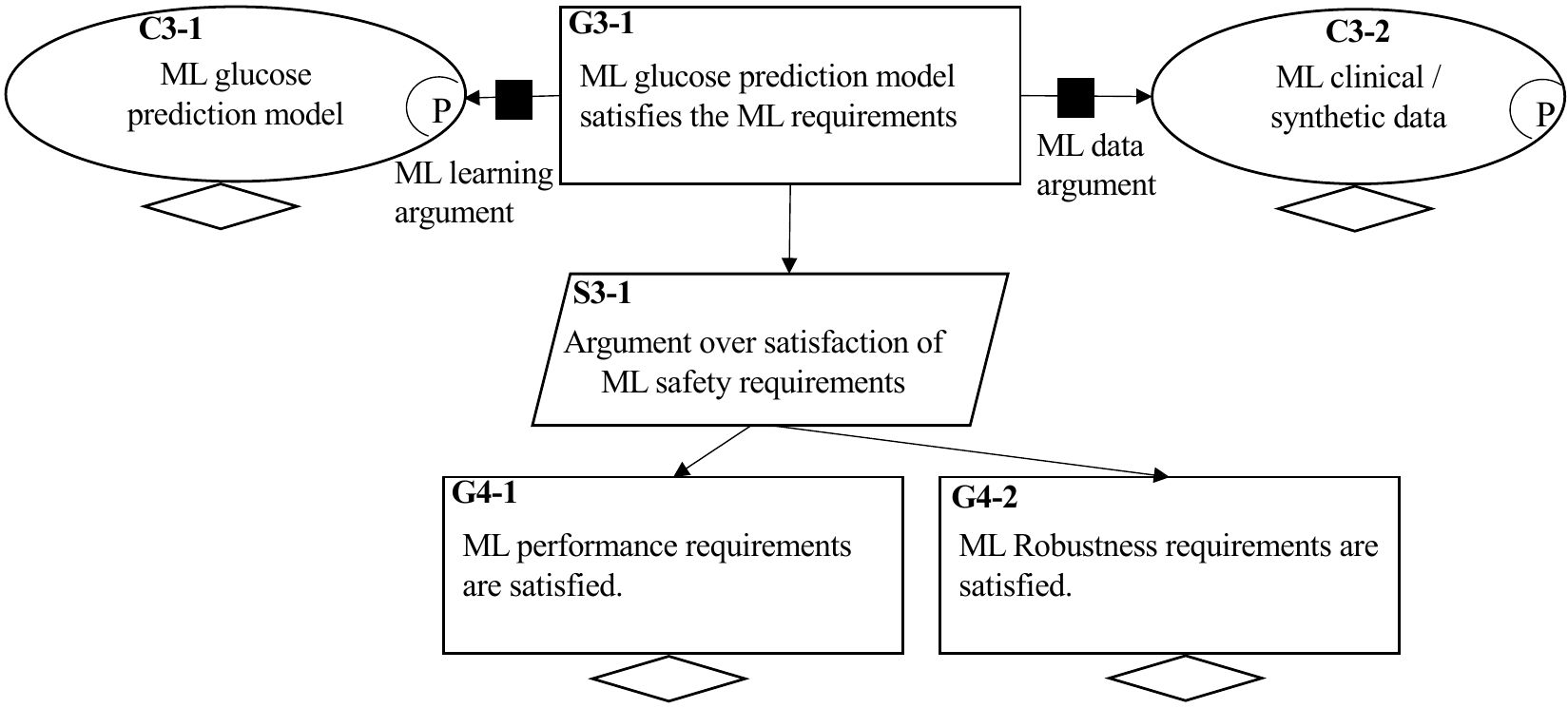}}
    \caption{Argument to ensure the sufficiency of the ML glucose prediction model. 
    }
    \label{fig::MLPartII}
    \vspace{-3em}
\end{figure}

The argument to ensure the sufficiency of the ML glucose prediction model is shown in Figure~\ref{fig::MLPartII}. The claim G3-1 is made in context C3-1 of the ML model created and context C3-2 of the ML data, 
respectively. 
ML data can include data from only an individual patient or a population of patients, and the ML model and its hyper-parameters are tuned based on collected data. 
The goal G3-1 is supported by goals G4-1 and G4-2 claiming that the ML model satisfies the performance and robustness requirements.  
Further assurance is also needed regarding the ML process and ML data used for development. The ML learning and ML data arguments provide arguments and evidence for the safety and effectiveness of the ML process and ML data. We provide an assurance case for the sufficiency of ML data in Section~\ref{subsec::data} and concrete evidence for both arguments in Section~\ref{sec::concEvid}. The links with the ML learning and data arguments are established using assurance claim points \cite{https://doi.org/10.48550/arxiv.2102.01564} (black squares), representing the points at which further assurance is required.

\vspace{-1em}
\subsection{Sufficiency of the ML Data} \label{subsec::data}
\begin{figure}
    \centering
    \fbox{\includegraphics[width=0.8\linewidth,keepaspectratio]{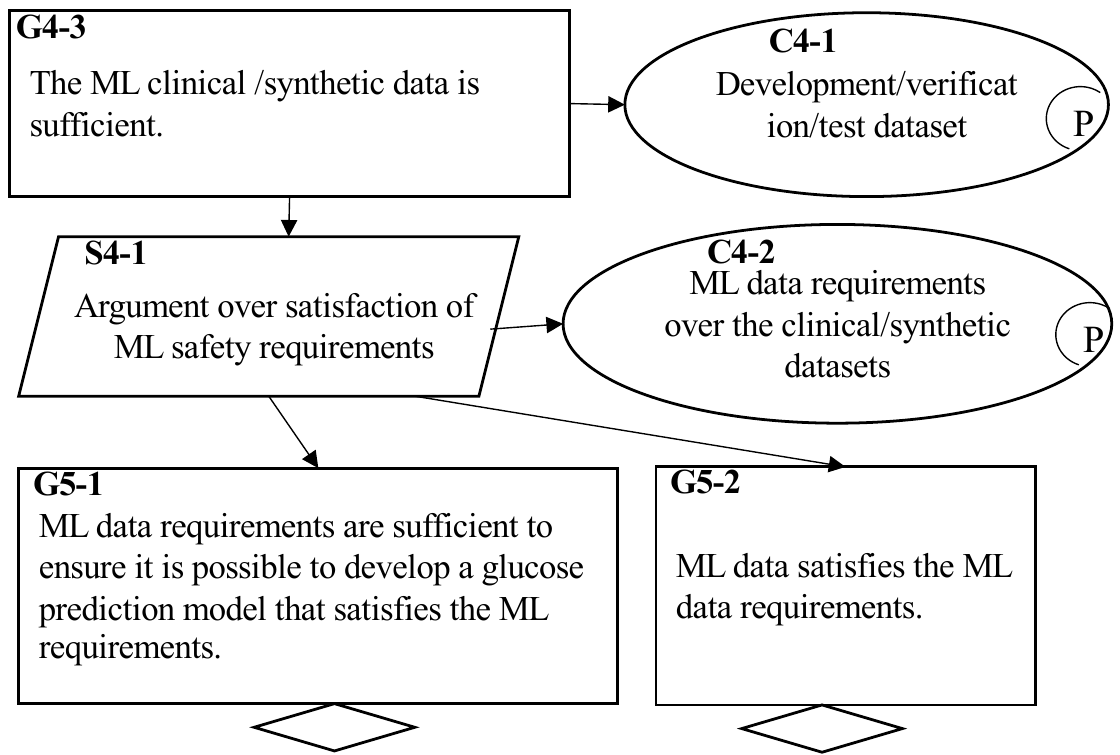}}
    \caption{Argument to ensure the sufficiency of the ML data. 
    }
    \label{fig::MLPartIII}
    \vspace{-2.5em}
\end{figure}

\vspace{-0.5em}
The argument to ensure the sufficiency of the ML data is shown in Figure~\ref{fig::MLPartIII}. Claim G4-3 
justifies that the data collected meet desiderata, including relevance, completeness, balance, and accuracy \cite{https://doi.org/10.48550/arxiv.2102.01564}, thus, the assurance  that the model trained on such data satisfies ML requirements increases. The first step to check the data against the desiderata is to provide a list of ML data requirements for  each desideratum. The sub-claim G5-1 assures that the list has sufficient ML data requirements, and the sub-claim G5-2 checks whether the data meet the ML data requirements. We enumerate the ML data requirements in Table~\ref{tab::dataReq}. Section~\ref{sec::concEvid} provides concrete evidence to support G5-2. In the following, we describe data requirements and their effects on the satisfaction of the performance and robustness requirements in support of G5-1.

\begin{table*}[]
\footnotesize
 \caption{Data Requirements in the ML lifecycle of an APS. 
    }
    \centering
    \begin{tabular}{m{1cm}|m{13cm}}
        \hline 
          \multicolumn{2}{|c|}{\textbf{Relevance}} \\
          \hline \hline
         DR.R1 & Each data sample shall assume sensor positioning which is representative of that used on the patients \\ \hline
         DR.R2 & The format of each data sample shall be representative of that captured using sensors deployed on the body \\ \hline
         DR.R3 & The type of each data sample (insulin) shall be representative of that used \\ \hline
         DR.R4 & Each data sample shall represent the diabetes type for which the system is developed \\ \hline
         DR.R5 & Each data sample shall represent the sex, age, and ethnicity of the persons for which the system is developed \\ \hline
         \multicolumn{2}{|c|}{\textbf{Completeness}} \\ \hline \hline
         DR.C1 & The data samples shall include examples with a sufficient range of meal carbs, different intraday meal intakes, and exercise \\ \hline
         DR.C2 & The data samples shall include examples with different sensor positioning \\ \hline
         DR.C3 & The data samples shall include examples with different ages and weights within the allowed ranges 
         \\ \hline
         DR.C4 & The data samples shall include patients with frequent hypoglycemic, hyperglycemic, and ketoacidosis problems \\ \hline
         DR.C5 & The data samples shall include the profile of patients during the day and night and illness \\ \hline
         \multicolumn{2}{|c|}{\textbf{Accuracy}} \\ \hline \hline
         DR.A1 & Each data sample shall assume sensor positioning which is representative of that used on the patients \\ \hline
         DR.A2 &  CGM sensor readings and pump infusions must be correctly recorded \\ \hline
         DR.A3 &  The total insulin delivered must be within the limit in each data sample \\ \hline
         \multicolumn{2}{|c|}{\textbf{Balance}} \\ \hline \hline
         DR.B1 & The datasets shall have a comparable number of samples for features \\ \hline
    \end{tabular}
    \label{tab::dataReq}
    \vspace{-2em}
\end{table*}

The requirements DR.R1 and DR.R2 concern the position of the CGM sensor on the patient's body and the format of the data captured by the sensor, respectively. A CGM sensor is 
worn on specific body areas. It should be placed around a fattier area of the body, i.e., the upper arm or abdomen for the adult and the abdomen or buttocks for kids. DR.R1 also relates to the accuracy desideratum (DR.A1), as the sensor position affects the accuracy of the sensor readings and, consequently, the accuracy of the BG prediction. The requirement DR.R3 refers to the type of insulin, i.e., rapid-acting, regular-acting, intermediate-acting, or long-acting, and even the brand of insulin. The APS controller may support a specific type of insulin, as different types have different absorption mechanisms.
The APS designed for adults may not be allowed to be used for kids or vice versa. As DR.R4 explains, a similar argument can be expressed for other characteristics such as gender, insulin type, etc., and relates to the relevance desideratum. Sex, age, and insulin type may affect the satisfaction of physiological and robustness properties.    

The APS controller should be able to safely adjust the insulin dosage in the face of uncertain events such as intraday meal intakes, exercise, and different values of meal carbohydrates. To support this, as explained by DR.C1, the datasets should include a sufficient range of examples in which the appropriate features refer to the mentioned events.  If the system is supposed to work for different positions of CGM sensor installments, as explained by DR.C2, sufficient examples regarding each position should be presented in the datasets. A similar requirement can be specified for weight and age. For instance, consider that the system is designed to work for people aged 14 to 60 who weigh between 20 and 120 pounds. The datasets should not only include sufficient samples with all allowed ages and weights, but also include samples with the combination of these features (DR.C3). 
DR.C4 is specified to ensure that the data samples include patients with frequent hypoglycemic, hyperglycemic, and ketoacidosis events. As specified by DR.C5, the data samples shall include the profile of patients during the day and night and even in sickness. Nighttime sleep and sickness impact metabolic regulation and endocrine release by the pancreas. Considering all the requirements above is crucial to satisfying physiological properties and ensuring robustness.

From the accuracy perspective, the CGM readings and the pump infusions not affected by a system failure must be correctly recorded, and the total amount of insulin delivered for each person be within the limit, as explained by DR.A2 and DR.A3, respectively. The only data requirement regarding the balance desideratum is that the number of samples for features should be comparable (DR.B1). For instance, the number of samples representing kids and adults should be comparable if the system is supposed to work for both categories of kids and adults. Notably, the dataset should include data from patients of different ages, sexes, weights, etc., if the context is defined for a population of patients. Hence, context C4-2 is annotated with P. Similarly, the size of the development, test, and verification datasets change according to the data collected and the model learned, which should be reflected in C4-1. So, C4-1 is also annotated with P.

%

%
%

\vspace{-1em}
\section{Concrete Evidence} \label{sec::concEvid}
\vspace{-0.5em}
In this section, we provide concrete evidence in support of ML learning argument, and claims G4-1, G4-2, and G5-2. 
We used the Simglucose simulator \cite{zhou2021design, Simglucose} to generate synthetic data for T1D patients and trained a Feed-Forward Neural Network (FFNN) to predict BG values. The model, contexts, and all properties in our experiments are  based on a population of patients. We performed our experiments on Ubuntu 20.04 with Intel Core i7, CPU 3.60GHz × 8, and 15.6 GiB memory. 

\begin{table*}[]
\footnotesize
    \centering
    \caption{ML Performance requirements. We use $BG_i, In_i$, and $M_i$ to denote BG, insulin, and meal intake, where $T^I=12$ and $T^O=6$. The superscript $I$ indicates the input and $O$ indicates the output of the network. We use $\Delta, \,\beta_1, \, \beta_2, \, \beta_3, \, \beta_4, \, \beta_5, \, \rho_1, \, \rho_2, \, \alpha$ to denote the thresholds in requirements.  $\Rightarrow$ denotes implication.  }
    \begin{tabular}{m{1.5cm}|m{11cm}}
        \hline 
          \multicolumn{2}{|c|}{\textbf{ML Performance Properties} } \\
          \hline \hline
         ML-RQ1.1 & $\bigwedge_{i=0}^{T^I-2} |BG_{i+1}^I-BG_i^I| \leq \Delta \Rightarrow \bigwedge_{j=0}^{T^O-2} |BG_{j+1}^O-BG_j^O| \leq \Delta$ \\ \hline
         ML-RQ1.2 &  $\bigvee_{i=0}^{T^I-1} M_i^I \geq \beta_1 \Rightarrow \bigvee_{j=0}^{T^O-1} BG_j^O \geq \rho_1$ \\ \hline
         ML-RQ1.3 & No available data  \\ \hline 
         ML-RQ1.4 &  $ In^I_0 \geq \beta_2 \Rightarrow \bigvee_{j=1}^{T^O-1} |BG^O_j-BG^O_0| \geq \alpha$ \\ \hline
         ML-RQ1.5 &  $ \bigvee_{i=0}^{T^I-1} M_i^I \geq \beta_3 \Rightarrow \bigvee_{j=1}^{T^O-1} |BG^O_j-BG^O_0| > 0$  \\ \hline        
         ML-RQ1.6 &   $In^I_0 \geq \beta_4 \Rightarrow 70 \leq BG_{T^O-1}^O \leq 180 \wedge \bigwedge_{j=0}^{j=T^O-2} (BG_j^O \leq 70 \vee BG_j^O \geq 180)$ \\ \hline   
         ML-RQ1.7 &  $In^I_0 \geq \beta_4 \Rightarrow 70 \leq BG_{T^O-1}^O \leq 180 \wedge \bigwedge_{j=0}^{j=T^O-2} (BG_j^O \leq 70 \vee BG_j^O \geq 180)$  \\ \hline  
         ML-RQ1.8 &  $\bigvee_{i=0}^{T^I-1} In_i^I \geq \beta_5 \Rightarrow \bigvee_{j=0}^{T^O-1} BG_j^O \leq \rho_2$ \\ \hline  
    \end{tabular}
    \label{tab::perRobReqML}
    \vspace{-2em}
\end{table*}

\textbf{ML Data} (\textit{Context C3-2}). 
Simglucose is a Python implementation of the FDA-approved UVA-Padova Simulator that employs a glucose-insulin meal model to simulate  30 virtual patients (ten adolescents, ten adults, and ten children). 
Using Simglucose, we emulated all patients  for 40 days and nights, where the BG and insulin values are provided every 5 minutes. 
Simglucose implements a basic basal-bolus controller and generates random meals for each patient, where the amount and the time of each meal are random numbers from pre-specified intervals. Each patient's data includes 11,521 entries, and each entry includes a set of features from which we use only BG, insulin, and meal data. We removed data of 4 adolescents, 1 adult, and 5 children from the dataset, since their data included negative BG values. 

\textbf{ML Model} (\textit{Context C3-1}). Our FFNN has three dense layers with 8, 8, and 6 neurons in each layer, respectively. 
It has 36 inputs, including BG, insulin, and meal intake of the patient for an hour (12 timesteps with 5 min intervals) and predicts BG values 30 minutes into the future (6 timesteps). We 
scale the inputs between 0 and 1. More details on the model are available at \cite{arxivpaper}. 

\textbf{Evidence for ML Learning Argument}. This argument grounds on the sufficiency of the iterative process to design and train the model. This process selects the model structure and appropriate values for the model parameters. 
We used the same number of neurons proposed in ~\cite{10.1007/978-3-319-99429-1_11,10.1145/3365365.3382210}. 
We tested the network with different neurons in each hidden layer and compared them using the root mean squared error (RMSE),  
which was very similar for those networks. We chose eight neurons in the first and second layers of the network, as network size affects verification complexity.
%
To make sure that the model does not overfit on data, we plotted training loss versus validation loss. 
We observed that validation loss decreases over the increasing number of epochs but, like training loss, becomes nearly fixed after a few epochs.

\textbf{Evidence for G4-1 and G4-2}. We need to ensure that the ML model meets each ML performance and robustness requirement. We used test-based verification to check ML-RQ1. We split the data into training and test data with a proportion of 80\% to 20\%, respectively (context C4-1), and calculated RMSE. 
We consider ML-RQ1 is satisfied if RMSE is less than a threshold (i.e., 12 mg/dL ~\cite{10.1007/978-3-319-99429-1_11}). The RMSE in our experiments is 3.03 mg/dL. We also used formal verification to check ML-RQ1.1 to ML-RQ1.8. We employed the DNNV framework ~\cite{10.1007/978-3-030-81685-8_6}, using which we compared the performance of different NN verifiers and selected Nnenum \cite{10.1007/978-3-030-76384-8_2}. The properties are specified using inputs and outputs of the network by constraining their ranges of values. Table~\ref{tab::perRobReqML} shows the mapping between the performance requirements allocated to the ML component (Table~\ref{tab::perRobReq}) and the requirements amenable to the ML implementation. We describe ML-RQ1.1 and ML-RQ1.2 as an example. In ML-RQ1.1, the difference between two consecutive BG values in the input and output is limited by $\Delta$. In ML-RQ1.2, we assume that if meal intake is larger than a value ($\beta_1$), BG will be greater than a value ($\rho_1$). We use an OR condition to indicate the timestep in which the meal is consumed is not relevant. The meal intake should be sufficiently large to assure us about its effect on the BG value.

%
\begin{table*}[]
\footnotesize
    \centering
    \caption{The properties checked on the FFNN of the glucose prediction. The third and forth columns indicate whether the property is satisfied over networks with 8,8,6 and 128,64,6 neurons. 
    We use $BG_i, In_i$, and $M_i$ to denote BG, insulin, and meal intake, where $i \in [1,12]$, $\alpha=0.006525$, and $\beta=[0,1]$. The superscript $I$ indicates the input and $O$ indicates the output. * denotes that the property was satisfied using another verifier (Marabou) as Nnenum raised error, and $\dagger$ shows that the property was satisfied after 16 days (using Marabou). The verification time for other requirements was fast enough. 
    }
    \begin{tabular}{m{4.6cm}|m{5.7cm}|m{0.75cm}|m{2cm}}
    \hline
     Property & Constraints & (8,8,6)  & (128,64,8) \\ \hline
     ML-RQ1.1 &   $ BG_i^I \in [130,180], In_i^I \in \beta, M_i^I \in \beta, \Delta=20$ & Satisfied & Nnenum Error* \\ \hline
    ML-RQ1.1 & $ BG_i^I \in [109,180], In_i^I \in \beta, M_i^I \in \beta, \Delta=20$ & Satisfied & Nnenum Error $\dagger$ \\ \hline
     ML-RQ1.8 ($In_1^I=5 \Rightarrow BG_6^O \leq 230$) & $ BG_i^I \in [212,230], In_{i\neq 1}^I = \alpha, M_i^I=0$ & Violated & Satisfied \\ \hline
    ML-RQ1.8 ($In_{12}^I=5 \Rightarrow BG_6^O \leq 230$) &  $  BG_i^I \in [211,220], In_{i\neq 12}^I = \alpha, M_i^I=0$ & Satisfied & Satisfied \\ \hline
    ML-RQ1.8 ( $In_{12}^I=5 \Rightarrow BG_6^O < 220$) & $ BG_i^I \in [212,222], In_{i\neq 12}^I = \alpha, M_i^I=0$ & Satisfied & Violated \\ \hline
     ML-RQ1.2 ( $ M_{12}^I=20 \Rightarrow BG_6^O > 210$)& $ BG_i^I \in [180,180], In_{i}^I = \alpha, M_{i \neq 12}^I=0$ & Violated & Satisfied \\ \hline
    ML-RQ1.2 ($ M_{12}^I=20 \Rightarrow BG_6^O > 200$)& $  BG_i^I \in [180,180], In_{i}^I = \alpha, M_{i \neq 12}^I=0$ & Satisfied & Satisfied \\ \hline
    ML-RQ1.2 ( $ M_{12}^I=20 \Rightarrow BG_6^O > 200$) & $ BG_i^I \in [180,183], In_{i}^I =\alpha , M_{i \neq 12}^I=0$ & Violated  & Satisfied \\ \hline
    \end{tabular}
    \label{tab::verResults}
     \vspace{-2em}
\end{table*}
%
To verify the properties, we first determined ranges of values based on the minimum and maximum values of the corresponding variables in the dataset. We also chose the thresholds based on our knowledge of the literature (e.g., we set $\Delta$, the constraint for max glucose
rise/drop over 5 min, to 40 based on ~\cite{10.1145/3365365.3382210}).  
As a result, all properties were violated. This confirms that learning complex body physiology in the presence of uncertain meal intake is difficult, and having high precision does not necessarily show the algorithm's correctness. Selecting the thresholds also  needs consulting with physicians and domain experts. So, we considered specific forms of the properties and selected the thresholds with try and test. We tried to increase the likelihood of property satisfaction by tightening ranges and thresholds. A part of our experiments are shown in Table~\ref{tab::verResults}. Besides, we checked the properties on a network with 128 and 64 neurons in layers one and two. We observed that a property satisfied on the first network is not necessarily satisfied on the other, and vice versa. These experiments confirm the need to instantiate AC according to the patient profiles or the population. Because the network structure as well as the thresholds and ranges of values may change based on data available for a patient or a population of patients. 


The robustness can be checked by measuring RMSE, given data of a virtual patient as the test data. The data does not include the exercise information (ML-RQ3).

\textbf{Evidence for G5-2}. 
Since we use synthetic data, the requirements DR.R1 (DR.A1), DR.R3, and DR.C2 in Table~\ref{tab::dataReq} are not applied to our dataset. Synthetic data generation was conducted via the Dexcom sensor, and this can serve as evidence to support DR.R2. Simglucose is a simulator to generate data of virtual patients with T1D, so DR.R4 is met. If the controller is used for all diabetic patients, DR.R5 and DR.C3 are violated since the data is generated for three subject groups of patients, excluding the elderly group and patients weighing more than 118 kg. We are also uncertain about sex and ethnicity. Simglucose generates random intraday meal intake. Thus, DR.C1 is partially met because the data does not include exercise information. Over the whole data, 0.14\% of the data samples are hyperglycemic, 0.11\% are hypoglycemic, and 99.75\% are in the glycemic range. Thus, DR.C4 is met, but the data balance, DR.B1, is violated. We are not certain about DR.C5. This requirement is met if the data generation model considers illness. Although the data is generated by the simulator, DR.A2 is satisfied because the simulator models both the sensor and pump. There are not equal numbers for three subject groups in the dataset
, which is another reason for the DR.B1 violation.








\vspace{-1em}

\section{Open Research Problems} \label{sec::openProblems}
\vspace{-0.5em}

Herein, based on our experience in developing safety assurance cases for learning-enabled MCPS, we outline several interesting open research problems. 

\noindent \textbullet \, We observed that the network structure influences the satisfaction or violation of a property. Undoubtedly, the training data, using which the weights in the network are calculated, also has an impact. How we can trace the violation of a property back to its origin?

\noindent \textbullet \, The difficulty of learning the patient's physiology solely from the training data may explain why several physiologically-based properties are violated. How can we enforce the ML model to satisfy the properties while it develops over the data?

\noindent \textbullet \, RNN is a very commonly used network for time series data. However, we are unaware of any RNN verifiers that can assess a broad range of properties (not just robustness) and are not specialized for specific applications. Also, the current FFNN verifiers do not support properties with complex structures like ours. How can we develop an RNN verifier functional for various properties?

In addition, addressing the following questions can improve the assurance case.

\noindent \textbullet \, How to develop adaptive safety AC for online learning models, e.g., where the datasets and consequently the learned model change during the system operation?

\noindent \textbullet \, How to develop quantitative measures to evaluate the confidence in a dynamic assurance case, via aggregating the uncertainty introduced by different evidence (e.g., from model training, testing, and verification) and reasoning about the sufficiency for assurance?

\noindent \textbullet \, How to build automated tool support for the development and review of safety AC for ML-enabled MCPS?

\vspace{-1em}

\section{Conclusion} \label{sec::conc}
\vspace{-0.8em}
In this paper, we presented a safety assurance case template for APS as a representative of learning-enabled MCPS. We focused on ensuring the safety and effectiveness of the ML-based APS controller. We first extracted the primary performance and robustness requirements allocated to the APS controller. Then we enumerated the requirements on the dataset and provided concrete evidence regarding ML and data requirements. In the future, we plan to continue this line of research and investigate the open  problems listed in Section~\ref{sec::openProblems}. 
\vspace{-1em}

\section*{Acknowledgment} 
\vspace{-0.8em}
This work was supported in part by the National Science Foundation (NSF) grants CCF-1942836, CCF-2131511, and CNS-2146295 and by the Commonwealth Cyber Initiative, an investment in the advancement of cyber R\&D, innovation, and workforce development.

\bibliography{software}

\appendix
\newpage
\section{Appendix}
\subsection{Glucose Prediction Model}
\begin{figure*}
    \centering
    \fbox{\includegraphics[width=0.9\textwidth,keepaspectratio]{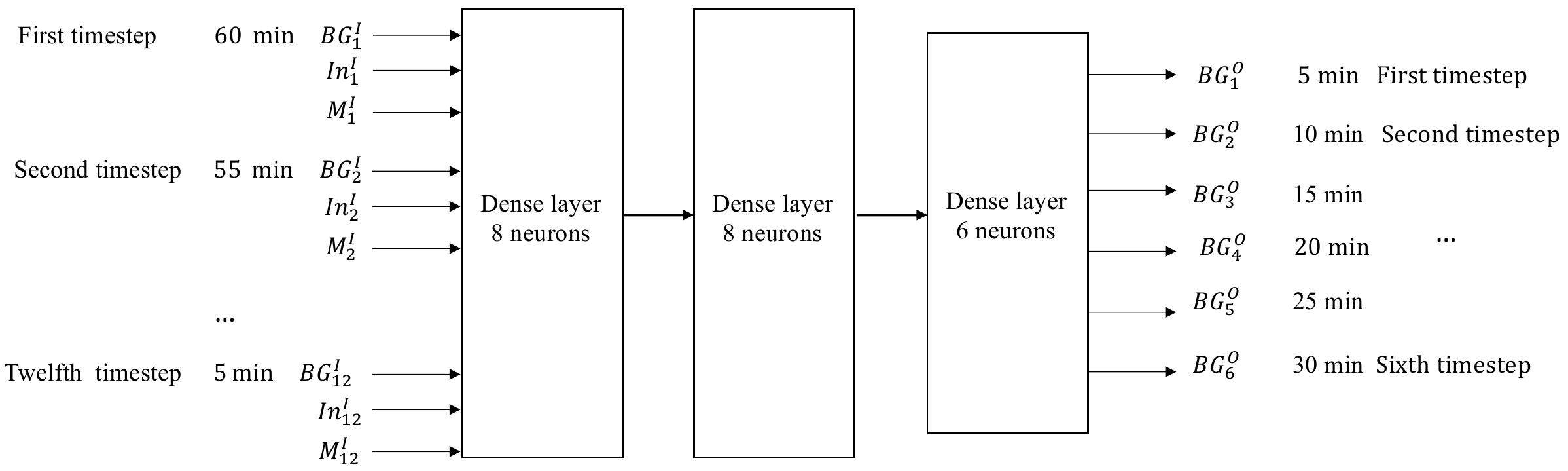}}
    \caption{The structure of our FFNN for glucose prediction with its inputs and outputs. The timesteps indicate the order in which the values are organized in the input and output sequences. The $t$ min in the input denotes $t$ minutes into the past, and $t$ min in the output denotes $t$ minutes into the future. $In$ denotes the insulin and $M$ denotes the meal.  }
    \label{fig:ffnnStructure}
\end{figure*}

we denote the structure of our FFNN along with its inputs and outputs in Figure~\ref{fig:ffnnStructure}. We use timesteps to indicate the order in which the values are organized in the input and output sequences. The $t$ min in the input denotes $t$ minutes into the past, and $t$ min in the output indicates $t$ minutes into the future.  We use MinMaxScalar to scale the inputs before feeding them to the network. We use the relu activation function and the adam optimizer to compile the model. 

\begin{figure*}
     \centering
     \begin{subfigure}[b]{0.47\textwidth}
         \centering
         \fbox{\includegraphics[width=\textwidth,keepaspectratio]{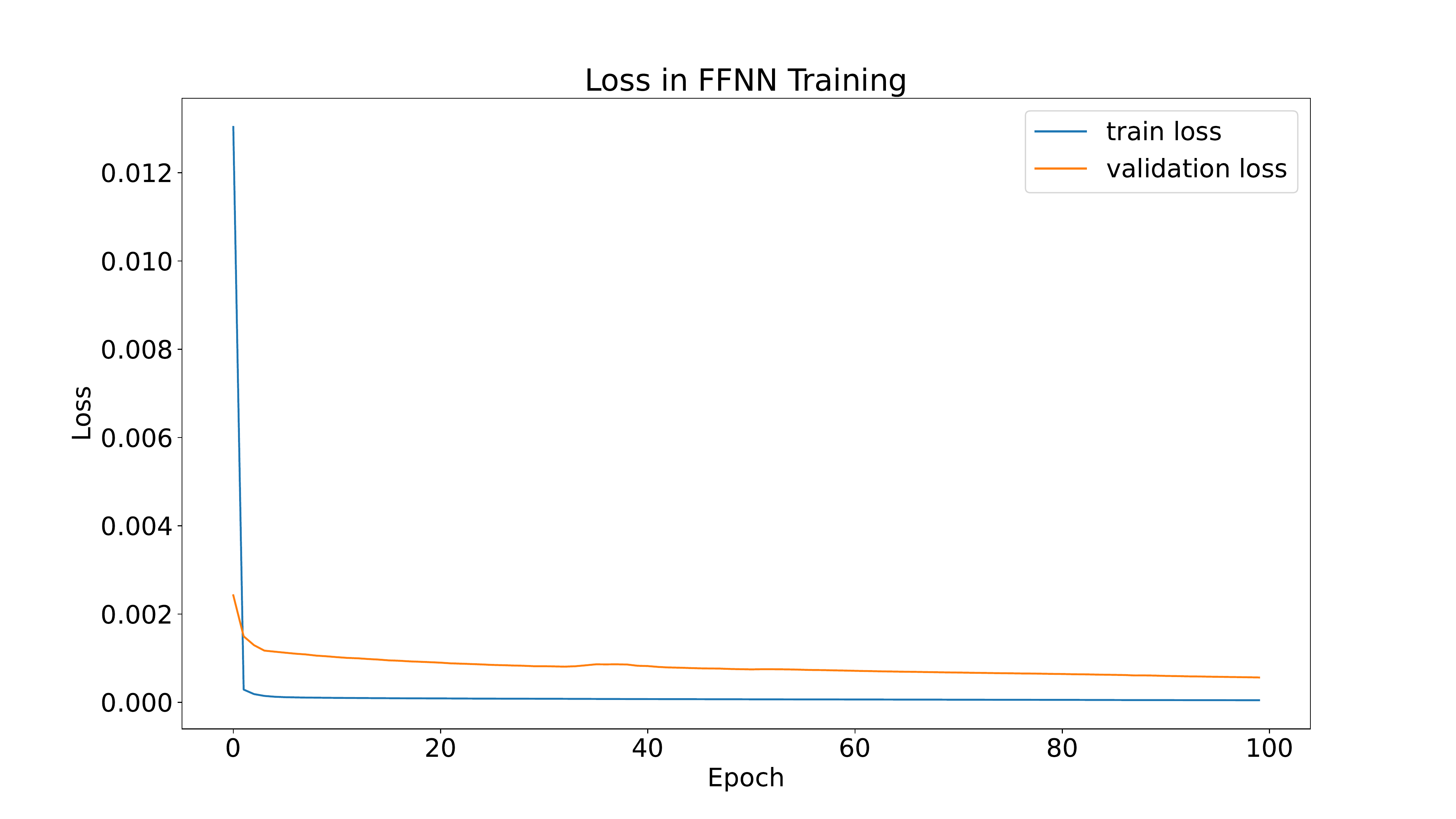}}
         \caption{}
         \label{fig::testVSvalidation}
     \end{subfigure}
     \hfill
     \begin{subfigure}[b]{0.47\textwidth}
         \centering
         \fbox{\includegraphics[width=\textwidth,keepaspectratio]{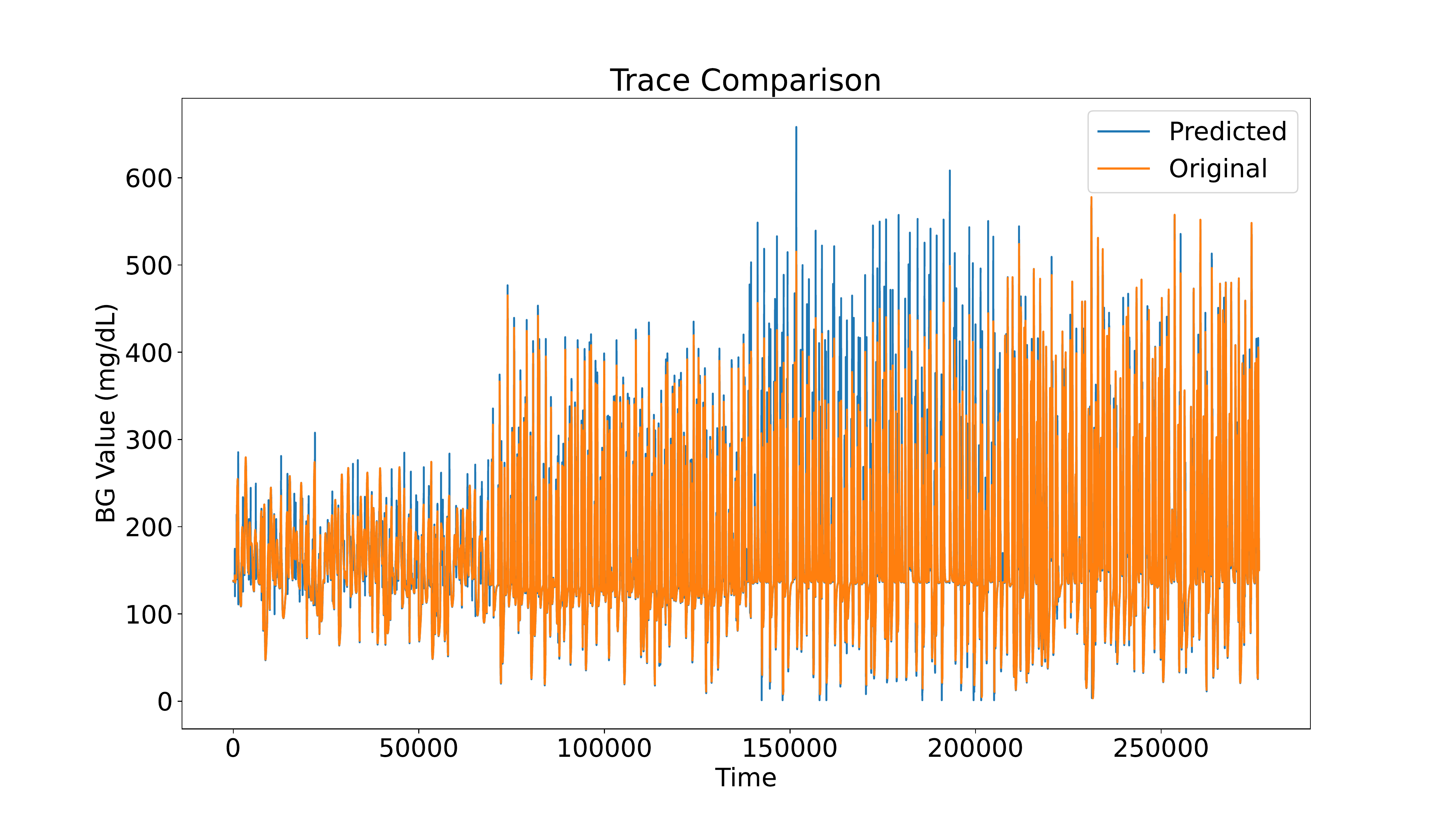}}
         \caption{}
         \label{fig::predVStrain}
     \end{subfigure}
        \caption{(a) Training loss versus validation loss for the FFNN with 8, 8, and 6 neurons in the first, second, and third layer, respectively, (b) The original and predicted BG values }
        \label{fig::predictionResults}
\end{figure*}

We show the training loss versus validation loss for our FFNN of the glucose prediction in Figure~\ref{fig::predictionResults}(a). The test data are compared with the predicted data in Figure~\ref{fig::predictionResults}(b). Our FFNN includes 8 and 8 neurons in the first and second layers, respectively. We provide RMSE for different combinations of neurons in two layers in Table~\ref{tab::mse}.

\begin{table}[]
    \caption{Root Mean Squared Error (RMSE) for different combinations of neurons in layers one and two}
    \centering
    \begin{tabular}{|c|c|}
        \hline 
          \textbf{Neurons in layer 1 and 2} & \textbf{RMSE}\\ \hline
         (8,8) & 3.03 \\ \hline
         (8,10) & 3.09  \\ \hline
         (8,20) & 3.11\\ \hline
         (8,64)& 3.13 \\ \hline
         (8,128)& 3.27 \\ \hline
         (8,200)& 3.19 \\ \hline
         (10,8)&  3.11 \\ \hline
         (20,8)& 3.14 \\ \hline
         (64,8)& 3.19 \\ \hline
         (128,8)& 3.19 \\ \hline
         (200,8) & 3.15\\ \hline
    \end{tabular}
    \label{tab::mse}
    \vspace{-2em}
\end{table}

\end{document}